\documentclass[10pt,twocolumn,letterpaper]{article}

\usepackage[pagenumbers]{iccv} 
\usepackage{comment}
\usepackage{multirow}
\usepackage{booktabs}
\usepackage{placeins} 
\usepackage{makecell}   
\usepackage{colortbl}   
\usepackage{float}
\usepackage{afterpage}
\usepackage{etoolbox,lipsum}
\usepackage{afterpage} 
\usepackage{graphicx}   
\usepackage{float}      
\DeclareUnicodeCharacter{064D}{}

\definecolor{iccvblue}{rgb}{0.21,0.49,0.74}
\usepackage[pagebackref,breaklinks,colorlinks,allcolors=iccvblue]{hyperref}
\usepackage{algorithm}
\usepackage{algpseudocode}
\def\paperID{11496} 
\def\confName{ICCV}
\def\confYear{2025}

\title{FedVSR: Towards Model-Agnostic Federated Learning in Video Super-Resolution}

\author{
    {\small \textbf{Ali Mollaahmadi Dehaghi}\textsuperscript{1},}
    {\small \textbf{Hossein KhademSohi}\textsuperscript{1},}
    {\small \textbf{Reza Razavi}\textsuperscript{2},}
    {\small \textbf{Steve Drew}\textsuperscript{1},}
    {\small \textbf{Mohammad Moshirpour}\textsuperscript{3}}
    \\
    {\small \textsuperscript{1} University of Calgary,}
    {\small \textsuperscript{2} Userful Corporation,}
    {\small \textsuperscript{3} University of California, Irvine}
    \\
    {\small \texttt{\{ali.mollaahmadidehag, hossein.khademsohi, steve.drew\}@ucalgary.ca},}\\
    {\small \texttt{reza.razavi@userful.com}, \texttt{mmoshirp@uci.edu}}
}

\begin{document}
\twocolumn[{%
\renewcommand\twocolumn[1][]{#1}%
\maketitle

}]


\begin{abstract}
Video super-resolution (VSR) aims to enhance low-resolution videos by leveraging both spatial and temporal information. While deep learning has led to impressive progress, it typically requires centralized data, which raises privacy concerns. Federated learning (FL) offers a privacy-friendly solution, but general FL frameworks often struggle with low-level vision tasks, resulting in blurry, low-quality outputs. To address this, we introduce \textit{FedVSR}, the first FL framework specifically designed for VSR. It is model-agnostic and stateless, and introduces a lightweight loss function based on the Discrete Wavelet Transform (DWT) to better preserve high-frequency details during local training. Additionally, a loss-aware aggregation strategy combines both DWT-based and task-specific losses to guide global updates effectively. Extensive experiments across multiple VSR models and datasets show that \textit{FedVSR} not only improves perceptual video quality (up to +0.89 dB PSNR, +0.0370 SSIM, $-0.0347$ LPIPS and 4.98 VMAF) but also achieves these gains with close to zero computation and communication overhead compared to its rivals. These results demonstrate \textit{FedVSR}’s potential to bridge the gap between privacy, efficiency, and perceptual quality, setting a new benchmark for federated learning in low-level vision tasks. The code is available at: \href{https://github.com/alimd94/FedVSR}{\textcolor{red}{https://github.com/alimd94/FedVSR}}

\end{abstract}    

\section{Introduction}

\begin{figure*}[t]
    \centering
    \footnotesize

    \begin{tabular}{@{}c@{\hspace{0.2cm}}c@{}}
        \begin{minipage}{7.5cm}
            \centering
            \includegraphics[width=\linewidth, height=6cm]{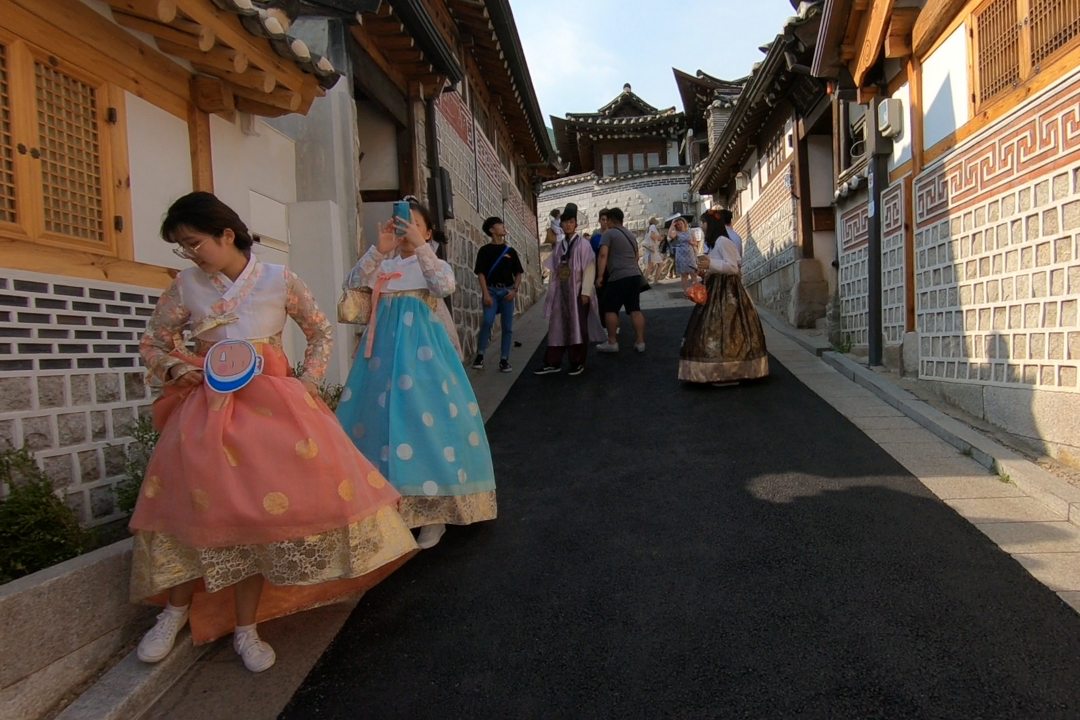}
            \\[3pt]
            \textbf{(a)} Frame 029, Clip 011, REDS\cite{Nah_2019_CVPR_Workshops}
        \end{minipage}
        &
        \begin{tabular}{@{}ccc@{}}
            \begin{minipage}{2.5cm}
                \centering
                \includegraphics[width=\linewidth, height=2.5cm]{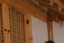}
                \\[0pt]
               \textbf{(b)} FedAvg\cite{mcmahan2017communication} \textcolor{red}{30.41}, \textcolor{blue}{0.8600}, \textcolor{green}{0.1777}
            \end{minipage} &
            \begin{minipage}{2.5cm}
                \centering
                \includegraphics[width=\linewidth, height=2.5cm]{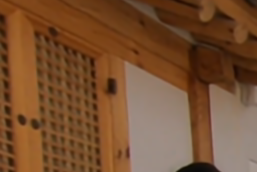}
                \\[0.0pt]
                \textbf{(c)} FedMedian \cite{pmlr-v80-yin18a} \textcolor{red}{28.89}, \textcolor{blue}{0.8224}, \textcolor{green}{0.2106}
            \end{minipage} &
            \begin{minipage}{2.5cm}
                \centering
                \includegraphics[width=\linewidth, height=2.5cm]{Images/rvrt_kinetics_avg_prox_029_011_c.png}
                \\[0pt]
                \textbf{(d)} FedProx\cite{li2020federated} \textcolor{red}{30.07}, \textcolor{blue}{0.8531}, \textcolor{green}{0.1851}
            \end{minipage} \\[1.5cm]
            \begin{minipage}{2.5cm}
                \centering
                \includegraphics[width=\linewidth, height=2.5cm]{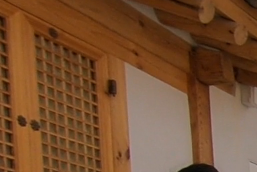}
                \\[0pt]
                \textbf{(e)} GT
                 \vspace{9pt}
            \end{minipage} &
            \begin{minipage}{2.5cm}
                \centering
                \includegraphics[width=\linewidth, height=2.5cm]{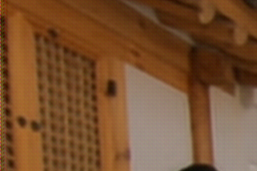}
                \\[0pt]
                \textbf{(f)} SCAFFOLD\cite{karimireddy2020scaffold} \textcolor{red}{26.78}, \textcolor{blue}{0.7148}, \textcolor{green}{0.2516}
            \end{minipage} &
            \begin{minipage}{2.5cm}
                \centering
                \includegraphics[width=\linewidth, height=2.5cm]{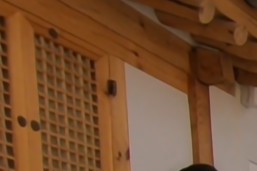}
                \\[0pt]
                \textbf{(g)} FedVSR (Ours) \textcolor{red}{32.00}, \textcolor{blue}{0.8889}, \textcolor{green}{0.1508}
            \end{minipage}
        \end{tabular}
    \end{tabular}

    \vspace{0.2cm}
    \captionsetup{justification=centering, singlelinecheck=false}
    \caption{Comparison of RVRT \cite{liang2022recurrent} output trained with different federated learning algorithms. General FL methods struggle to reconstruct fine details and textures. PSNR, SSIM, and LPIPS metrics are shown in \textcolor{red}{red}, \textcolor{blue}{blue}, and \textcolor{green}{green} respectively.}

    \label{fig:comparison}
\end{figure*}

\label{sec:intro}
Video super-resolution (VSR) is an advanced computational process that reconstructs high-resolution (HR) video sequences from low-resolution (LR) counterparts, leveraging both spatial and temporal information \cite{liu2022learning}. Unlike single-image super-resolution (SISR), which relies solely on intra-frame details, VSR exploits temporal correlations across consecutive frames to enhance the quality and sharpness of the output. This capability is crucial in real-world applications such as surveillance \cite{deshmukh2019fractional}, video streaming \cite{kim2020neural}, and medical imaging \cite{robinson2017new, peng2020saint}, where visual fidelity and temporal coherence are mission-critical. In practice, VSR must also cope with degradations such as down-sampling \cite{zhang2024real}, noise \cite{zhou2024upscale}, blur \cite{morris2024dabit}, and compression artifacts \cite{Dehaghi_2025_WACV}. However, modern deep-learning pipelines address these degradations with data-hungry networks trained on centrally collected data, but centralization itself raises privacy, regulatory, and bandwidth concerns, especially when the raw videos contain sensitive or personally identifiable information.

Federated learning (FL) provides a principled remedy to these concerns \cite{mcmahan2017communication}. In numerous Video Super-Resolution (VSR) scenarios, such as remote patient monitoring, personal phone footage analysis, and confidential surveillance, video data captured on edge devices is inherently privacy-sensitive. Training on this data centrally is often prohibited by stringent regulations such as GDPR or CCPA \cite{https://doi.org/10.1049/ise2/5536763}, which forbid uploading and aggregating raw footage containing personally identifiable information. Privacy is the core motivator for FL in VSR. This decentralization also solves the massive logistical challenge of video data size; for instance, attempting to upload even a short 10-second 8K video for training can easily exceed 10 GB \cite{10.1145/3587819.3592560}. Unlike conventional distributed methods that rely on frequent synchronization, FL enables selected participants to perform multiple local updates before their models are aggregated into a global one. This approach minimizes communication overhead and accommodates the limited computational resources of edge devices. These combined benefits make FL a logical and effective foundation for VSR systems that face such concerns.

Despite significant advancements in FL methods, most research has focused on high-level or downstream visual tasks such as image classification \cite{karimireddy2020scaffold, li2020federated, wang2020tackling}, facial recognition \cite{liu2021feddg}, and segmentation \cite{miao2023fedseg, liu2021feddg}. However, their application in super-resolution (SR) remains largely unexplored. Existing studies \cite{moser2024federated, yang2025fedsr} found that directly applying general FL techniques to low-level vision tasks like SR results in poor-quality reconstructions with blurred textures. This failure is compounded in VSR, where data heterogeneity (diverse video content and quality) and communication constraints severely degrade the model's ability to capture and synchronize rich spatial and temporal high-frequency details across frames, a prerequisite for high-fidelity video output.
Motivated by these limitations, we propose \textit{FedVSR}, a novel framework for VSR under FL. \textit{FedVSR} is model-agnostic, meaning it treats underlying VSR models as black boxes and works by modifying only the local training loss and the server aggregation strategy. This enables compatibility with diverse VSR architectures without requiring internal model modifications or access to intermediate layers, which is crucial for scalability and flexibility in a real-world federated deployment. The framework consists of two key components. The first component is a novel, frequency-aware training phase enhanced by a lightweight 3D Discrete Wavelet Transform (DWT)-based loss. By operating in the frequency domain, this loss forces clients to prioritize the spatial and temporal high-frequency details essential for VSR that are typically lost due to FL's limited communication budget. Crucially, we demonstrate that this loss is specific to the FL environment, as its direct application in centralized VSR leads to performance degradation (see Section \ref{sec:abl_loss_center}). The second is an innovative adaptive loss aggregation mechanism that explicitly uses client training quality to mitigate the detrimental effects of data heterogeneity and noisy clients. While high-frequency preservation is important for all SR tasks, it is especially crucial in VSR to maintain temporal coherence across frames. Federated settings further amplify this challenge due to data heterogeneity and limited communication. By combining its two components, \textit{FedVSR} enhances training and aggregation robustness while remaining scalable, state-free, and practical for privacy-sensitive, resource-constrained environments. In summary, the main contributions of this paper are as follows:

\begin{itemize}[topsep=0pt, partopsep=0pt, itemsep=0pt, parsep=0pt]
\item We design a two-pronged solution to the Federated VSR challenge: (1) a novel, frequency-aware loss using 3D DWT to preserve critical spatial-temporal high-frequency details during decentralized training, and (2) an adaptive loss-aware aggregation method that dynamically adjusts client weights to combat data heterogeneity. This combination represents the first holistic framework designed to bridge the gap between VSR quality and FL privacy.

\item Our approach makes no architecture-specific assumptions and does not require storing past weights, making it stateless and model-agnostic. This design reduces memory usage, avoids privacy risks, and eliminates synchronization overhead in federated learning on edge devices.

\item Extensive experiments validate \textit{FedVSR}'s effectiveness and robustness among existing approaches and paving the way for privacy-preserving low-level vision in the FL setting.

\end{itemize}

\begin{figure*}[ht]
    \centering
    \includegraphics[width=\textwidth]{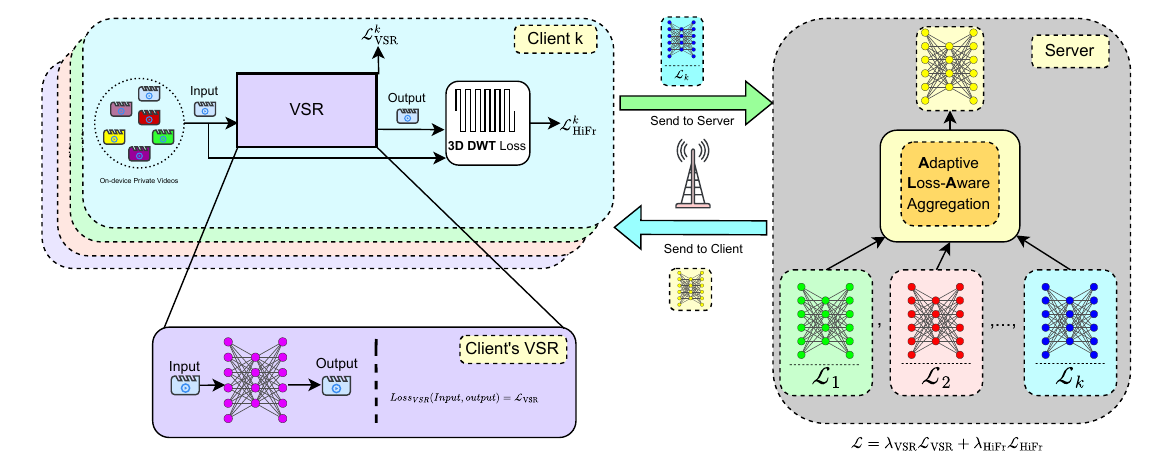}
  \caption{Overview of the proposed \textit{FedVSR} framework. Each client computes a model-agnostic VSR update augmented with a DWT-based high-frequency loss. Clients also track the average local loss , which is used for loss-aware weighted aggregation at the server. The global model is iteratively refined while maintaining model-agnostic and stateless properties.}

  \label{fig:teaser}
\end{figure*}

\section{Related Works}
\label{sec:relatedWorks}
In this section, we begin by discussing the most significant contributions in general FL and VSR. We then explore studies that specifically focus on applying FL to the SR task.
\subsection{Video Super-Resolution}
Early VSR methods that applied SISR techniques frame-by-frame neglected the temporal context, resulting in temporal inconsistencies, loss of detail, and undesired smoothing in the high resolution output \cite{liu2022video}. VSR leverages multiple temporally correlated low-resolution frames within a video sequence to generate high-resolution output. Unlike SISR, VSR must account for both spatial and temporal dependencies across frames, making it a highly complex, non-linear, multi-dimensional problem.

Recent advancements in VSR models have introduced various techniques to enhance performance. VRT \cite{liang2024vrt} enables parallel processing of long video sequences using a mutual attention mechanism for feature extraction, alignment, and fusion. RVRT \cite{liang2022recurrent} combines parallel and recurrent approaches, employing guided deformable attention for precise clip alignment. IART \cite{xu2024enhancing} uses implicit resampling with local cross-attention to preserve frequency details while reducing spatial distortions. DiQP \cite{Dehaghi_2025_WACV}, the first model for 8K videos, employs Denoising Diffusion to counteract compression artifacts while maintaining robustness across varying levels.

\subsection{Federated Learning}
Federated learning (FL) is a decentralized paradigm that enables multiple clients to collaboratively train models while preserving data privacy. It has been widely applied in domains such as healthcare, finance, and the Internet of Things (IoT), where data security and regulatory compliance are critical \cite{aledhari2020federated, chai2024survey, zhao2024federated}. A foundational approach is FedAvg \cite{mcmahan2017communication}, which follows a four-step iterative process: (1) the server distributes a global model to all clients, (2) each client trains its local model on its own data, (3) locally updated models are sent back to the server, and (4) the server aggregates these updates to refine the global model for the next round.

To improve FL performance, several methods focus on optimizing local updates. FedProx \cite{li2020federated} adds a proximal term to maintain consistency between local and global models. SCAFFOLD \cite{karimireddy2020scaffold} introduces control variates for both clients and server, while FedNova \cite{wang2020tackling} employs normalized averaging to address objective inconsistencies. More recently, MOON \cite{li2021model} leverages contrastive learning to assess similarity among different model representations.

Although these methods enhance convergence and stability in classification tasks, their reliance on class-specific assumptions limits applicability to low-level, regression-based tasks such as VSR. For instance, MOON, FedDyn \cite{acar2021federatedlearningbaseddynamic}, FedDisco \cite{pmlr-v202-ye23f}, and FedLAW \cite{pmlr-v202-li23s} depend on logits, class-wise statistics, or server-side data proxies—assumptions that do not translate to VSR or SISR scenarios.

\subsection{Federated Learning in Image Super Resolution}
Based on our review of the literature, FL has not yet been explored for VSR; the works discussed below are currently the only FL-based approaches applied to SISR. For Image-based low-level vision tasks, FedMRI \cite{feng2022specificity} introduces a federated learning framework for MRI reconstruction using a shared encoder for generalized features and client-specific decoders for personalization. FedPR \cite{feng2023learning} leverages pre-trained models and prompt learning with minimal trainable parameters to improve image quality. p²FedSR \cite{yang2024exploiting} enhances  personalized privacy-preserving SR through client-specific degradation priors and selective model aggregation, while FedSR \cite{yang2025fedsr} employs detail-assisted contrastive loss for feature alignment and a hierarchical aggregation policy for layer-wise model integration.

Although these approaches represent significant progress in FL for low-level vision, they are restricted to SISR rather than VSR. Moreover, methods such as FedSR, p²FedSR, and FedMRI are closely tied to specific model architectures (shallow||deep||upsampler), limiting applicability to a narrow range of SR frameworks. They also rely on stateful training: clients must retain auxiliary information such as degradation priors, feature statistics, or historical checkpoints, which introduces non-trivial storage, computation, and synchronization overhead. Such constraints hinder scalability and deployment, especially in heterogeneous or resource-limited environments.

In this paper, we propose \textit{FedVSR}, a model-agnostic and stateless framework that allows any VSR model to be seamlessly incorporated without structural modifications or client-side state management. This design ensures easy extensibility, offering greater flexibility and scalability than prior work, and making \textit{FedVSR} practical for real-world deployment where lightweight, generalizable solutions are essential.

\section{Overview of \textit{FedVSR}}
\label{sec:overview}

\textit{FedVSR} integrates Video Super-Resolution (VSR) with Federated Learning (FL), enabling collaborative enhancement of video resolution across distributed clients while preserving data privacy. This section provides a concise overview of the underlying VSR and FL problems and their formulations.

\subsection{Video Super-Resolution (VSR)}
\label{sec:vsr}

VSR aims to enhance video resolution by leveraging temporal information across multiple low-resolution (LR) frames, exploiting inter-frame dependencies that single-image super-resolution (SISR) methods cannot utilize:

Let \( F_i \in \mathbb{R}^{H \times W \times 3} \) denote the \( i \)-th frame in an LR video sequence, and \( \hat{F}_i \in \mathbb{R}^{sH \times sW \times 3} \) its high-resolution (HR) counterpart, where \( s \) is the upscaling factor. Define a temporal window of \( 2N+1 \) HR frames \( \{\hat{I}_j\}_{j=i-N}^{i+N} \) centered around \( \hat{I}_i \). The LR frames are generated via a degradation process:

\begin{equation}
I_i = \phi(\hat{I}_i, \{\hat{I}_j\}_{j=i-N}^{i+N}; \theta_{\alpha}),
\end{equation}

where \( \phi(\cdot; \cdot) \) models degradation factors such as blur, noise, and motion. A more explicit formulation is:

\begin{equation}
I_j = D B E_{i \to j} (\hat{I}_i) + n_j,
\end{equation}

with \( D \) as the down-sampling operator, \( B \) the blur effect, \( E_{i \to j} \) a warping function for motion compensation, and \( n_j \) image noise. VSR seeks to reconstruct HR frames:

\begin{equation}
\tilde{I}_i = \phi^{-1}(I_i, \{I_j\}_{j=i-N}^{i+N}; \theta_{\beta}),
\end{equation}

where \( \tilde{I}_i \) approximates the ground truth \( \hat{I}_i \) \cite{ulyanov2018deep, liu2022video}. This is an ill-posed problem due to the non-injective nature of the degradation function.

\subsection{Federated Learning (FL)}
\label{sec:fl}

FL allows multiple clients to collaboratively train a shared model without exchanging raw data, reducing privacy risks and communication overhead. The global optimization problem is:

\begin{equation}
\min_{\mathbf{w}} F(\mathbf{w}) = \sum_{k=1}^{K} p_k F_k(\mathbf{w}),
\end{equation}

where \( K \) is the number of clients, \( p_k \) is the relative weight of client \( k \), \( F_k(\mathbf{w}) \) the local loss, and \( \mathbf{w} \) the global model parameters. Clients update locally via:

\begin{equation}
\mathbf{w}_k^{t+1} = \mathbf{w}^t - \eta \nabla F_k(\mathbf{w}^t),
\end{equation}

and the server aggregates updates with weighted averaging:

\begin{equation}
\mathbf{w}^{t+1} = \sum_{k=1}^{K} p_k \mathbf{w}_k^{t+1}.
\end{equation}

Client participation can vary due to data heterogeneity, resource limits, or network instability, and simple dataset-size weighting may not capture each client's true impact \cite{zhan2021survey, caldarola2023window, mcmahan2017communication, qi2024model}. \textit{FedVSR} incorporates hybrid adaptive loss-based weighting, where clients with lower local loss contribute more effectively to the global model, improving training stability and performance. See Section \ref{sec:loss}

\subsection{\textit{FedVSR}: Integrating VSR with FL}

\textit{FedVSR} combines these paradigms by distributing the VSR training process across multiple clients. Each client improves HR frame reconstruction locally while preserving its private video data. The server aggregates model updates using adaptive weights to optimize global performance. This approach maintains temporal coherence in videos, leverages multi-client data diversity, and preserve privacy to address the challenges of centralized VSR training.

\section{Proposed \textit{FedVSR} Framework}
VSR remains a challenging problem due to its ill-posed nature. Severe degradations amplify artifacts, making it difficult to balance detail enhancement and noise suppression. To improve generalizability, diverse training data and large batch sizes are needed, but this significantly increases computational costs \cite{chan2022investigating}. FL adds further complexity to VSR due to real-world constraints \cite{fan2024survey} and the nature of the task. In heterogeneous environments, differences in client data distributions lead to biased updates, poor generalization \cite{caldarola2022improving}, and inefficient model merging. Gradient mismatches \cite{xiao2020averaging} and parameter inconsistencies slow convergence \cite{crawshaw2023federated}. Our design is a two-pronged approach that enhances client training via a novel loss term and improves server robustness through adaptive aggregation. The overall architecture of \textit{FedVSR} framework is illustrated in Fig.~\ref{fig:teaser}

\label{sec:Methodology}
\subsection{High-Frequency Loss for Federated VSR}
Selecting a proper loss function as objective for VSR model is crucial and determine the overall output's performance \cite{vasu2018analyzing,song2022dual} and as depicted in Fig.~\ref{fig:comparison}. It is evident that existing FL algorithms has problems in capturing and generating details in the output. Since we aim to avoid adding computational overhead to the client by using adversarial or perceptual loss, and because our objective is to keep the model untouched, ensuring \textit{FedVSR} remains model-agnostic, we introduce an additional term to the overall loss of model improve performance.
\subsubsection{\textbf{3D DWT for High-Frequency Preservation}}
One common failure in training VSR models is producing overly smooth outputs, meaning the trained model fails to capture high-frequency information. As a result, details and textures are missing from the output. Since the models we used are among the top-performing ones in a centralized setting, this suggests that the failure to restore high-frequency details is specifically related to training under the FL paradigm. To alleviate this, we leverage the 3D Discrete Wavelet Transform (DWT). While 2D DWT has shown effectiveness in SISR \cite{8237449, zhao2024wavelet}, we apply the 3D variant to capture temporal high-frequency details across frames in addition to spatial ones. This is essential for maintaining temporal coherence in VSR output, a problem standard FL aggregation fails to address effectively. 

DWT is a multi-resolution signal decomposition technique that represents a signal in terms of approximation and detail coefficients at different scales. For our implementation, we select the separable Haar wavelet. While more sophisticated wavelets offer better decorrelation, Haar's inherent simplicity and use of short integer filters result in minimal computational overhead during the local update step. This is a critical design choice for FL, as it aligns with our goal of introducing a lightweight, non-intensive loss term that does not significantly burden the resource-constrained client devices. For a 3D signal (such as a volumetric image or a video clip), DWT decomposes the signal into eight sub-bands of different frequencies and orientations:

\begin{itemize}
    \item Low-Frequency Approximation Coefficients \( F_{i}^{LLL} \) (low-pass filtered along all three dimensions).
    \item High-Frequency Coefficients with single high-pass filtering:
    \begin{itemize}
        \item \( F_{i}^{LLH} \) (low-pass depth and height, high-pass width),
        \item \( F_{i}^{LHL} \) (low-pass depth and width, high-pass height),
        \item \( F_{i}^{HLL} \) (low-pass height and width, high-pass depth).
    \end{itemize}
    \item High-Frequency Coefficients with two high-pass filterings:
    \begin{itemize}
        \item \( F_{i}^{LHH} \) (low-pass depth, high-pass height and width),
        \item \( F_{i}^{HLH} \) (low-pass height, high-pass depth and width),
        \item \( F_{i}^{HHL} \) (low-pass width, high-pass depth and height).
    \end{itemize}
    \item Full High-Frequency Coefficients \( F_{i}^{HHH} \) (high-pass in all three dimensions).
\end{itemize}

Using the separable Haar wavelet, the 1D filters are defined as:
\begin{itemize}
    \item Low-pass filter \( h = \frac{1}{\sqrt{2}} [1, 1] \),
    \item High-pass filter \( g = \frac{1}{\sqrt{2}} [1, -1] \).
\end{itemize}

The coefficients in the eight sub-bands are computed by successive filtering along the three dimensions (depth, height, width). For example:
\begin{align}
    F_{i}^{LLL} &= ((F_i * h^T) * h) * h
\end{align}
\begin{align}
    F_{i}^{LLH} &= ((F_i * h^T) * h) * g
\end{align}
\begin{align}
    F_{i}^{LHL} &= ((F_i * h^T) * g) * h
\end{align}
\begin{align}
    F_{i}^{LHH} &= ((F_i * h^T) * g) * g
\end{align}
\begin{align}
    F_{i}^{HLL} &= ((F_i * g^T) * h) * h
\end{align}
\begin{align}
    F_{i}^{HLH} &= ((F_i * g^T) * h) * g
\end{align}
\begin{align}
    F_{i}^{HHL} &= ((F_i * g^T) * g) * h
\end{align}
\begin{align}
    F_{i}^{HHH} &= ((F_i * g^T) * g) * g.
\end{align}

Similar to the 2D case, the low-frequency component \( F_{i}^{LLL} \) captures the global structure and coarse content of the input, while the seven high-frequency sub-bands capture edges, textures, and structural variations along different orientations \cite{priyadharsini2018wavelet}. 

For our loss design, we focus on the high-frequency sub-bands and ignore the low-frequency approximation.
The high-frequency helper loss is then computed using the Charbonnier loss \cite{charbonnier1994two}:

\begin{equation}
    \mathcal{L}_{\mathrm{HiFr}}(\bar{F}, \hat{F})
    = \mathbb{E}\!\left[
        \sqrt{\big( H_{\text{HF}}(\bar{F}) - H_{\text{HF}}(\hat{F}) \big)^{2} + \epsilon^{2}}
    \right]
\end{equation}

where \(H_{\text{HF}}(\cdot)\) denotes the concatenation of those seven high frequency sub-bands \(\{LLH, LHL, LHH, HLL, HLH, HHL, HHH\}\) obtained by the 3D DWT from the predicted \(\bar{F}\) and ground truth \(\hat{F}\) frame , 
\(\epsilon > 0\) is the Charbonnier constant, and 
\(\mathbb{E}[\cdot]\) indicates the mean over all elements (batch, channels, temporal, and spatial).

Finally, the total training objective combines the standard VSR loss with the DWT-based high-frequency loss:
\begin{equation}
    \mathcal{L}_{\text{total}} = \lambda_{\text{VSR}} \mathcal{L}_{\text{VSR}} + \lambda_{\text{HiFr}} \mathcal{L}_{\text{HiFr}}
\end{equation}

\subsection{Adaptive Loss-Aware Aggregation}
\label{sec:loss}
To further enhance the model's performance, we employ a hybrid aggregation strategy that dynamically balances uniform and loss-based weighting.

In general VSR research, the standard task loss (e.g., L1 or L2) is already widely accepted as a reliable pixel-wise measure of reconstruction quality \cite{liu2022video}. This provides a crucial distinction from federated classification tasks. The total local loss ($\mathcal{L}_{\text{total}}$) used in $\textit{FedVSR}$ maintains this reliability, being calculated as the combination of the standard VSR loss and our novel $\mathcal{L}_{\text{HiFr}}$ high-frequency loss. Our $\mathcal{L}_{\text{HiFr}}$ term acts as an enhancement, sensitizing $\mathcal{L}_{\text{total}}$ to the high-frequency details often lost through averaging, thus making the total loss an even better measure of performance. Unlike high-level tasks such as classification where a high loss might signal valuable Non-IID data \cite{cmes.2023.027226} , a high value for $\mathcal{L}_{\text{total}}$ in VSR directly signifies a failure to reconstruct high-fidelity details and temporal coherence, the fundamental objective. Therefore, clients with significantly higher $\mathcal{L}_{\text{total}}$ pixel-wise losses genuinely contribute lower-quality model updates, making prioritization justifiable.

This aggregation strategy, inspired by MKL-SGD \cite{shah2020choosing}, improves robustness by prioritizing updates that achieve superior local performance. Our method first computes the average local loss ($\mathcal{L}_{\text{total}}$) for each client after training, which is then transmitted to the server as a single scalar value alongside the model weights. 


\subsubsection{\textbf{Deriving Loss-Based Prioritization}}

Rather than using inverse loss weights directly, we normalize them and apply an adaptive step parameter to create the loss-based distribution ($l_i$), which naturally emphasizes clients with lower losses.

\subsubsection{\textbf{Hybrid Weighting for Robustness}}

A key design feature is that we do not rely on $l_i$ alone. The aggregation is governed by a convex combination of $l_i$ and the uniform distribution ($u_i$). This $\mathbf{u_i}$ component ensures that all clients contribute a baseline minimum update, thereby preventing the complete marginalization of updates derived from challenging, yet strategically important, diverse video data.

\subsubsection{\textbf{Thresholded Adaptation via Hellinger Distance}}

To avoid overreacting to minor variations in client performance, we quantify the divergence between the uniform ($u_i$) and loss-based ($l_i$) distributions using the Hellinger distance \cite{Hellinger1909}. This distance serves as a mixing coefficient ($m$). By applying a soft threshold ($\tau$) to this distance, the loss-based prioritization is only significantly activated when there is a large, meaningful divergence in client update quality. If losses are similar, the aggregation gradually defaults to uniform averaging.

Finally, the client's weight ($w_i$) in the global update is the convex combination of these two distributions, which is then applied to the model parameters during aggregation. This approach allows the aggregation process to adaptively leverage loss-based prioritization when needed, while safeguarding against the unfair penalization of diverse data, thereby improving both robustness and generalization. The overall algorithm is summarized in Algorithm \ref{alg:fl_dwt}.

\begin{algorithm}[ht]
    \small
    \caption{Our Method (\textit{FedVSR})}
    \label{alg:fl_dwt}
    \begin{algorithmic}[1]
        \Require Number of communication rounds \( T \), Number of clients \( N \), Learning rate \( \eta \), Adaptive step \( \alpha \), Decay rate \( \gamma \)
        \Ensure Global model weights \( W^T \)
        
        \State \textbf{Server Initialization:} Initialize global model \( W^0 \)
        
        \For{\( t = 1, \dots, T \)}
            \State Server selects a subset of clients randomly \( S_t \subseteq \{1, \dots, N\} \)
            
            \For{\textbf{each} client \( i \in S_t \) in parallel}
                \State Client downloads latest global model \( W^t \)
                \State Perform local training for \( E \) epochs with learning rate \( \eta \)
                
                \State Compute local losses:
                \State \quad \( \mathcal{L}_{\text{VSR}} \) (task-specific loss)
                \State \quad \( \mathcal{L}_{\text{HiFr}} \) (high-frequency loss)

                \State Compute total loss: 
                \[
                    \mathcal{L}_i = \lambda_{\text{VSR}} \mathcal{L}_{\text{VSR}} + \lambda_{\text{HiFr}} \mathcal{L}_{\text{HiFr}}
                \]

                \State Update local model:
                \[
                    W_i^{t+1} = W^t - \eta \nabla F_i(W^t)
                \]

                \State Return \( W_i^{t+1} \) and loss value \( \mathcal{L}_i \) to server
            \EndFor
            
            \State \textbf{Server Aggregation:}
            \State Compute uniform weights \( u_i = \frac{1}{|S_t|} \)
            \State Compute inverse-loss weights \( I_i = \frac{1}{\mathcal{L}_i} \), apply adaptive step \( I'_i = I_i^{\alpha} \)
            \State Normalize loss-based weights \( l_i = \frac{I'_i}{\sum_{j \in S_t} I'_j} \)
            \State Compute Hellinger distance:
            \[
                H = \sqrt{\tfrac{1}{2} \sum_{i \in S_t} (\sqrt{u_i} - \sqrt{l_i})^2}
            \]
            \State Compute mixing coefficient:
            \[
                m = 
                \begin{cases} 
                0 & H < \tau \\
                \tfrac{H - \tau}{1 - \tau} & H \ge \tau
                \end{cases}
            \]
            \State Final aggregation weights: \( w_i = (1 - m) u_i + m l_i \)
            \State Update global model: 
            \[
                W^{t+1} = \sum_{i \in S_t} w_i \cdot W_i^{t+1}
            \]
            \State Apply Decay Rate to Adaptive Step \(\alpha = \alpha^{1 - \frac{t}{T}}\)
        \EndFor
        \State \Return Global model \( W^T \)
    \end{algorithmic}
\end{algorithm}

\section{Experiments}
\label{sec:Results}

\begin{table*}[htbp!]

\setlength{\tabcolsep}{7pt}
\centering
\begin{tabular}{|c|c|c|cccc|cccc|}
\hline
\multirow{3}{*}{\textbf{Model}} &
\multirow{3}{*}{\makecell{\textbf{Params}\\(M)}} &
\multirow{3}{*}{\textbf{Algorithm}} &
\multicolumn{8}{c|}{\rule{0pt}{2.5ex}\textbf{Less Heterogeneous}\rule[-1.5ex]{0pt}{3ex}} \\
\cline{4-11}
& & &
\multicolumn{4}{c|}{\textbf{REDS} \cite{Nah_2019_CVPR_Workshops}} &
\multicolumn{4}{c|}{\textbf{Vid4} \cite{6549107}} \\
& & &
{\scriptsize PSNR$\uparrow$} & {\scriptsize SSIM$\uparrow$} & {\scriptsize LPIPS$\downarrow$} & {\scriptsize VMAF$\uparrow$} &
{\scriptsize PSNR$\uparrow$} & {\scriptsize SSIM$\uparrow$} & {\scriptsize LPIPS$\downarrow$} & {\scriptsize VMAF$\uparrow$} \\
\hline

\multirow{5}{*}{\makecell{\textbf{VRT}\\\cite{liang2024vrt}}} & \multirow{5}{*}{35.6}
& FedAvg \cite{mcmahan2017communication} & 29.71 & 0.8522 & 0.1916 & 96.59 & 24.88 & 0.7680 & 0.2446 & 81.62 \\
& & FedProx \cite{li2020federated} & 28.82 & 0.8281 & 0.2133 & 92.95 & 24.18 & 0.7341 & 0.2668 & 76.03 \\
& & SCAFFOLD \cite{karimireddy2020scaffold} & 28.59 & 0.8181 & 0.2228 & 91.45 & 23.99 & 0.7182 & 0.2737 & 73.15 \\
& & FedMedian \cite{pmlr-v80-yin18a} & 27.89 & 0.7999 & 0.2379 & 87.46 & 23.59 & 0.6997 & 0.2963 & 70.62 \\
& & \textbf{\textit{FedVSR} (Ours)} & \textbf{29.96} & \textbf{0.8595} & \textbf{0.1807} & \textbf{97.47} & \textbf{24.97} & \textbf{0.7745} & \textbf{0.2284} & \textbf{83.07} \\
\hline

\multirow{5}{*}{\makecell{\textbf{RVRT}\\\cite{liang2022recurrent}}} & \multirow{5}{*}{10.8}
& FedAvg \cite{mcmahan2017communication} & 31.30 & 0.8877 & 0.1536 & 97.97 & 25.96 & 0.8156 & 0.1974 & 86.05 \\
& & FedProx \cite{li2020federated} & 30.09 & 0.8618 & 0.1899 & 95.65 & 25.20 & 0.7775 & 0.2330 & 82.01 \\
& & SCAFFOLD \cite{karimireddy2020scaffold} & 29.63 & 0.7967 & 0.2101 & 94.71 & 24.79 & 0.7317 & 0.2407 & 81.01 \\
& & FedMedian \cite{pmlr-v80-yin18a} & 29.76 & 0.8444 & 0.1931 & 94.79 & 24.88 & 0.7602 & 0.2404 & 80.11 \\
& & \textbf{\textit{FedVSR} (Ours)} & \textbf{31.64} & \textbf{0.8956} & \textbf{0.1377} & \textbf{98.94} & \textbf{26.10} & \textbf{0.8207} & \textbf{0.1802} & \textbf{87.37} \\
\hline

\multirow{5}{*}{\makecell{\textbf{IART}\\\cite{xu2024enhancing}}} & \multirow{5}{*}{13.4}
& FedAvg \cite{mcmahan2017communication} & 30.56 & 0.8729 & 0.1695 & 97.84 & 25.43 & 0.7949 & 0.2118 & 84.34 \\
& & FedProx \cite{li2020federated} & 29.65 & 0.8500 & 0.1683 & 96.14 & 24.72 & 0.7599 & 0.2307 & 80.65 \\
& & SCAFFOLD \cite{karimireddy2020scaffold} & 30.29 & 0.8678 & 0.1692 & 97.47 & 25.18 & 0.7857 & 0.2084 & 83.38 \\
& & FedMedian \cite{pmlr-v80-yin18a} & 29.17 & 0.8265 & 0.1960 & 95.55 & 24.40 & 0.7447 & 0.2302 & 78.50 \\
& & \textbf{\textit{FedVSR} (Ours)} & \textbf{30.98} & \textbf{0.8845} & \textbf{0.1473} & \textbf{98.45} & \textbf{25.59} & \textbf{0.8077} & \textbf{0.1899} & \textbf{86.12} \\
\hline
\end{tabular}
\caption{Comparison of PSNR~($\uparrow$), SSIM~($\uparrow$), LPIPS~($\downarrow$), and VMAF~($\uparrow$) for different FL algorithms under \textbf{less heterogeneous} data distributions across VSR models and datasets.}
\label{tab:fl_comparison_less}
\end{table*}

\begin{table*}[htbp!]

\setlength{\tabcolsep}{7pt}
\centering
\begin{tabular}{|c|c|c|cccc|cccc|}
\hline
\multirow{3}{*}{\textbf{Model}} &
\multirow{3}{*}{\makecell{\textbf{Params}\\(M)}} &
\multirow{3}{*}{\textbf{Algorithm}} &
\multicolumn{8}{c|}{\rule{0pt}{2.5ex}\textbf{More Heterogeneous}\rule[-1.5ex]{0pt}{3ex}} \\
\cline{4-11}
& & &
\multicolumn{4}{c|}{\textbf{REDS} \cite{Nah_2019_CVPR_Workshops}} &
\multicolumn{4}{c|}{\textbf{Vid4} \cite{6549107}} \\
& & &
{\scriptsize PSNR$\uparrow$} & {\scriptsize SSIM$\uparrow$} & {\scriptsize LPIPS$\downarrow$} & {\scriptsize VMAF$\uparrow$} &
{\scriptsize PSNR$\uparrow$} & {\scriptsize SSIM$\uparrow$} & {\scriptsize LPIPS$\downarrow$} & {\scriptsize VMAF$\uparrow$} \\
\hline

\multirow{5}{*}{\makecell{\textbf{VRT}\\\cite{liang2024vrt}}} & \multirow{5}{*}{35.6}
& FedAvg \cite{mcmahan2017communication} & 28.85 & 0.8282 & 0.2200 & 91.81 & 24.60 & 0.7516 & 0.2557 & 77.52 \\
& & FedProx \cite{li2020federated} & 28.64 & 0.8236 & 0.2222 & 88.04 & 24.10 & 0.7256 & 0.2713 & 72.16 \\
& & SCAFFOLD \cite{karimireddy2020scaffold} & 28.17 & 0.8074 & 0.2318 & 87.83 & 24.04 & 0.7208 & 0.2739 & 71.85 \\
& & FedMedian \cite{pmlr-v80-yin18a} & 27.68 & 0.7761 & 0.2429 & 85.72 & 23.77 & 0.7001 & 0.2890 & 69.69 \\
& & \textbf{\textit{FedVSR} (Ours)} & \textbf{29.12} & \textbf{0.8371} & \textbf{0.2010} & \textbf{94.37} & \textbf{24.78} & \textbf{0.7646} & \textbf{0.2378} & \textbf{80.10} \\
\hline

\multirow{5}{*}{\makecell{\textbf{RVRT}\\\cite{liang2022recurrent}}} & \multirow{5}{*}{10.8}
& FedAvg \cite{mcmahan2017communication} & 29.89 & 0.8578 & 0.1867 & 96.31 & 25.67 & 0.8040 & 0.2065 & 84.13 \\
& & FedProx \cite{li2020federated} & 29.59 & 0.8526 & 0.1943 & 94.71 & 25.02 & 0.7730 & 0.2353 & 82.01 \\
& & SCAFFOLD \cite{karimireddy2020scaffold} & 26.71 & 0.7142 & 0.2590 & 72.38 & 23.08 & 0.6441 & 0.3018 & 57.54 \\
& & FedMedian \cite{pmlr-v80-yin18a} & 28.85 & 0.8202 & 0.2119 & 87.84 & 24.51 & 0.7419 & 0.2499 & 74.73 \\
& & \textbf{\textit{FedVSR} (Ours)} & \textbf{31.00} & \textbf{0.8844} & \textbf{0.1492} & \textbf{97.45} & \textbf{25.95} & \textbf{0.8197} & \textbf{0.1823} & \textbf{85.85} \\
\hline

\multirow{5}{*}{\makecell{\textbf{IART}\\\cite{xu2024enhancing}}} & \multirow{5}{*}{13.4}
& FedAvg \cite{mcmahan2017communication} & 29.52 & 0.8493 & 0.1909 & 95.99 & 25.04 & 0.7774 & 0.2272 & 81.71 \\
& & FedProx \cite{li2020federated} & 29.30 & 0.8444 & 0.1958 & 94.16 & 24.42 & 0.7505 & 0.2293 & 77.42 \\
& & SCAFFOLD \cite{karimireddy2020scaffold} & 29.26 & 0.8413 & 0.1992 & 94.16 & 24.85 & 0.7690 & 0.2278 & 79.40 \\
& & FedMedian \cite{pmlr-v80-yin18a} & 28.80 & 0.8280 & 0.2330 & 88.90 & 23.88 & 0.7350 & 0.2555 & 73.99 \\
& & \textbf{\textit{FedVSR} (Ours)} & \textbf{29.85} & \textbf{0.8568} & \textbf{0.1787} & \textbf{96.84} & \textbf{25.31} & \textbf{0.7914} & \textbf{0.2118} & \textbf{83.18} \\
\hline
\end{tabular}
\caption{Comparison of PSNR~($\uparrow$), SSIM~($\uparrow$), LPIPS~($\downarrow$), and VMAF~($\uparrow$) for different FL algorithms under \textbf{more heterogeneous} data distributions across VSR models and datasets.}
\label{tab:fl_comparison_more}
\end{table*}

\subsection{setup}
\subsubsection{\textbf{Dataset:}} \label{sec:data}
We designed two distinct training environments to comprehensively evaluate the robustness and generalization of $\textit{FedVSR}$ under varying levels of data heterogeneity, utilizing two foundational video datasets: REDS \cite{Nah_2019_CVPR_Workshops} and Kinetics-400 \cite{kay2017kineticshumanactionvideo}. The REDS dataset is high-quality and professionally recorded, designed specifically for video restoration tasks, offering high content diversity but remaining relatively controlled, with sequences at $720 \times 1280$ resolution. In contrast, Kinetics-400,dataset primarily used for human action recognition, is a massive dataset sourced from unique amateur YouTube videos, whose content is highly realistic and exhibits significant camera shake, lighting variations, and low control, making it ideal for simulating highly diverse, uncontrolled, edge-device data.

For the Less Heterogeneous Scenario, we utilize the training sequences of REDS, splitting the data randomly ($\mathbf{IID}$-like) among clients to simulate a less diverse video environment. For the More Heterogeneous Scenario, we utilize a custom split of Kinetics-400 to simulate a strong $\mathbf{Non\text{-}IID}$ environment. Critically, because VSR does not use labels and developing an natural metric for video heterogeneity would be out of scope, we exploit the existing human action labels (40 classes selected) of Kinetics-400 solely to partition the video clips into Non-IID client silos, maximizing the distribution shift between clients. Only $720 \times 1280$ videos are selected from Kinetics-400 for format consistency with REDS.

To assess the generalization capability of the models trained under both scenarios, we use two standard external test sets: the REDS Test set and the Vid4 dataset \cite{6549107}. The models trained on both the REDS and Kinetics-400 splits are consistently evaluated against both benchmarks. The reason for selecting these specific datasets for testing is to strictly adhere to VSR community standards for better comparison of performance and measurement, since they are established benchmarks.

\subsubsection{\textbf{Implementation Details:}} \label{sec:mainexp}
To maintain our objective of developing a model-agnostic framework, we did not alter the official parameters of the baseline VSR models. Instead, we used their recommended settings for training. We created 40 clients, with participation of 10\% in both experimental setups. In the more heterogeneous setup, each client received videos from a single class; for example, one client had only "archery" videos, while another had only "making pizza" videos. In contrast, in the less heterogeneous setup, the six videos were randomly assigned to each client. We set the total number of global rounds to 100, with each client performing one local round (epoch) per global round. After completing their local training, these clients sent their updated weights to the server for aggregation. All experiments were conducted using bi-cubic (BI) degradation.
\subsubsection{\textbf{Baselines:}}
For our baseline, we use a combination of FedAvg \cite{mcmahan2017communication}, FedProx \cite{li2020federated}, FedMedian \cite{pmlr-v80-yin18a}, and SCAFFOLD \cite{karimireddy2020scaffold} from the FL side, along with VRT \cite{liang2024vrt}, RVRT \cite{liang2022recurrent}, and IART \cite{xu2024enhancing} from VSR. We selected these FL methods as they are widely accepted and frequently used to benchmark core algorithmic performance in recent literature. 

\subsubsection{\textbf{Evaluation Metrics:}}The evaluation metrics used for the experiments are Peak Signal-to-Noise Ratio (PSNR) and Structural Similarity Index (SSIM) and Learned Perceptual Image Patch Similarity (LPIPS) and Video Multimethod Assessment Fusion (VMAF).

\subsubsection{\textbf{Reproducibility:}} \label{sec:repro} To ensure our experiments are reproducible, we fixed the random seed, which doesn't affect client selection, model initialization, etc. We used four NVIDIA A6000 GPUs for training, with each model under a federated learning algorithm taking an average 46 hours to train. For the baselines alone, this consumed 1102 hours (2 experiments × 3 VSR models × 4 Fl algorithms × 46 hours per run).

\subsection{Results and Comparison}
\label{sec:res}
\subsubsection{\textbf{Quantitative Results:}} As shown in Tables ~\ref{tab:fl_comparison_less} and ~\ref{tab:fl_comparison_more}, our proposed \textit{FedVSR} algorithm consistently outperforms other federated learning (FL) algorithms across all tested models and datasets. On average, \textit{FedVSR} surpasses existing FL methods by a margin of 0.82 dB in PSNR and 0.0337 in SSIM and 0.0334 in LPIPS and 3.62 in VMAF in less heterogeneous settings, while in more heterogeneous settings, the margins are 0.96 dB in PSNR and 0.0404 in SSIM and 0.0359 in LPIPS and 6.34 in VMAF, demonstrating its robustness under challenging conditions. Among the evaluated models, RVRT benefits the most from \textit{FedVSR}, achieving the highest performance margin. In contrast, IART shows the lowest overall improvement but still maintains a competitive edge over the baseline FL methods.

\begin{figure*}[t]
    \centering
    \includegraphics[width=0.99\linewidth]{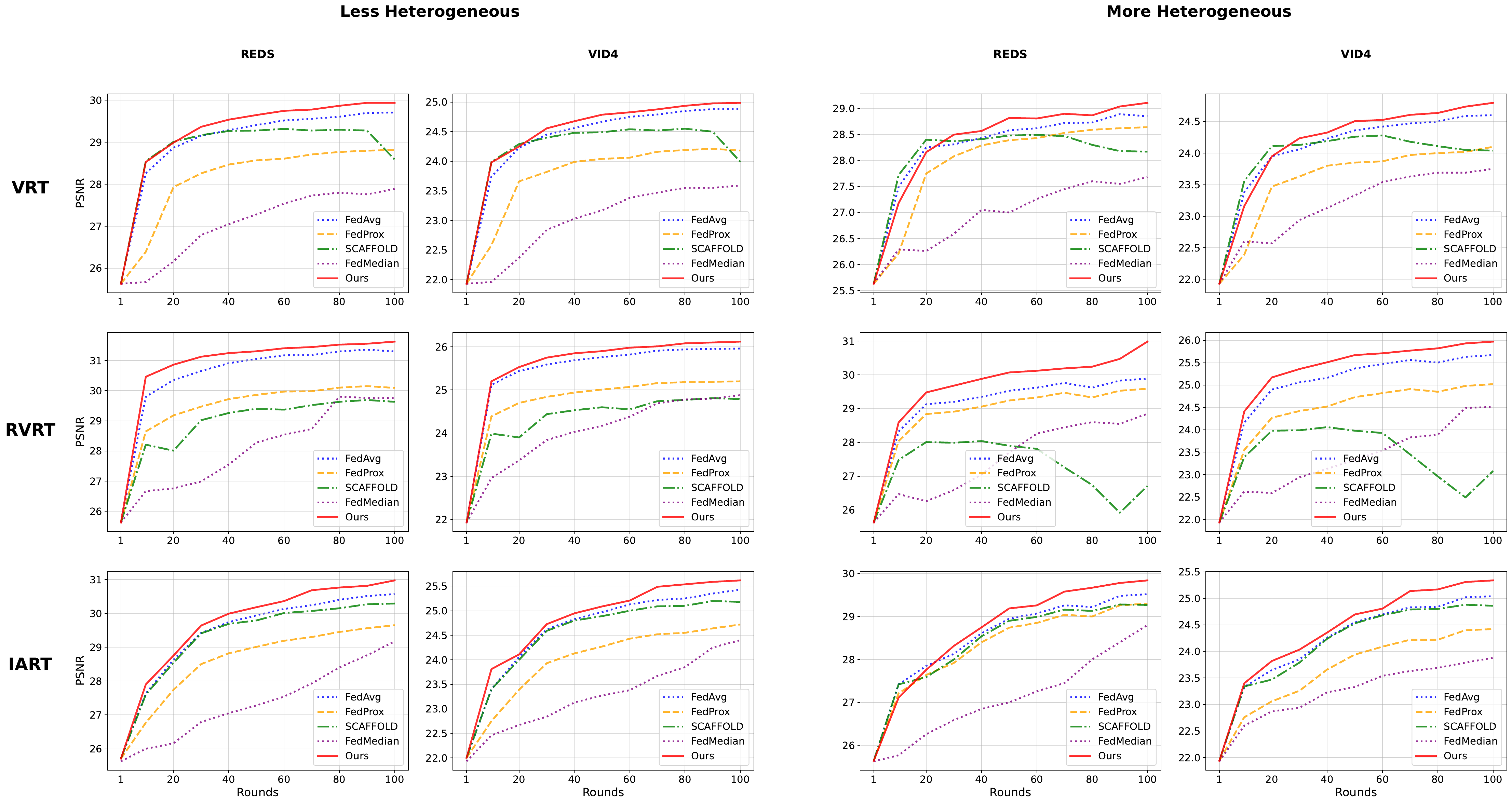} 
    \caption{PSNR across different rounds for various test sets under different settings for VRT \cite{liang2024vrt}, RVRT \cite{liang2022recurrent}, and IART \cite{xu2024enhancing}.}
    \label{fig:psnr_rounds}
\end{figure*}

The observed performance trends in our experiments align with existing findings on FedProx, SCAFFOLD and FedMedian further validating our results. As shown in Fig.~\ref{fig:psnr_rounds}, \textit{FedVSR} consistently achieves superior PSNR values across different VSR models and datasets, outperforming FL baselines particularly in later training rounds. This aligns with prior research on FedProx, which highlights its failure to reach the stationary points of the global optimization objective, even under homogeneous data distributions. The study suggests that FedProx’s limiting points may remain far from the true minimizer of the global empirical risk \cite{su2022nonparametricviewfedavgfedprox}, which could explain its suboptimal performance in our experiments. Although FedProx may still generalize well by aggregating local data effectively in certain conditions, its weaker convergence properties become apparent in more heterogeneous scenarios. We also observe that FedProx is highly sensitive to the \( \mu \) value during training. The original paper suggests \( \{0.001, 0.01, 0.1, 1\} \) as candidate values and finds that \( \mu = 1 \) performs better in most cases. However, we notice that for any \( \mu \) greater than 0.001, the VSR model struggles significantly with learning. This suggests that the relationship is more complex and requires further investigation and analysis.

Similarly, our findings regarding SCAFFOLD’s performance in highly heterogeneous settings support previous research showing that it is very sensitive to differences in client data \cite{iot3020016}. We observed significant drops and stagnation in our experiments, particularly with the RVRT model, which matches reports that even small changes in local data batches can cause large swings in performance \cite{iot3020016}. SCAFFOLD struggles to train effectively in such diverse scenarios, leading to slower or stalled convergence, which is reflected in our results \ref{fig:psnr_rounds} where its PSNR fluctuates in highly heterogeneous scenarios. The increased variance in SCAFFOLD’s performance under skewed data distributions further supports our findings that it fails to adapt smoothly across different client datasets, reinforcing its limitations in federated VSR tasks. On top of this, SCAFFOLD requires roughly twice the communication overhead of other algorithms due to its control variates, which further limits its practicality. See Section \ref{sec:overhead}

Although FedMedian demonstrates better stability than SCAFFOLD in our experiments, it clearly suffers from a lower convergence rate, which is consistent with prior findings. Prior work \cite{pmlr-v206-zhu23b} shows that the median-based update is less statistically efficient than mean or FedAvg when each client has a limited local dataset. Similarly, it is shown that coordinate-wise median becomes increasingly inefficient as the model grows in size or complexity, making it sub-optimal in high-dimensional settings. Additional empirical studies \cite{MICHALAKOPOULOS20254223} note that FedMedian often converges more slowly because it lacks explicit regularization and tends to produce more conservative updates. 

Overall, these existing studies provide strong theoretical support for our experimental results. In contrast, \textit{FedVSR}’s convergence, stability, and robustness across varying levels of data heterogeneity highlight its effectiveness in overcoming these limitations.

\subsubsection{\textbf{Qualitative Results:}} As showen in Fig. ~\ref{fig:comparison} clearly demonstrate the limitations of FedAvg, FedProx, and SCAFFOLD in accurately reconstructing fine details. While these methods reduce some pixelation from the LQ input, they all fail to recover the textures and sharp edges present in the ground-truth. FedAvg produces overly smooth results, blurring important structural details such as the wooden lattice and frame edges. FedProx exhibits similar blurring, struggling to restore high-frequency textures. SCAFFOLD, not only fails to produce a sharp reconstruction but also introduces noticeable artifacts, distorting the texture of the wooden lattice and making the output appear unnatural. In contrast, \textit{FedVSR} delivers significantly sharper and more realistic reconstructions, clearly outperforming the other approaches.

\subsection{Extended Evaluation and Test Cases}
\subsubsection{\textbf{Additional Baseline: FedSR vs \textit{FedVSR}}} As discussed in Section~\ref{sec:relatedWorks}, existing FL methods for the SISR task were designed for a different setting, providing detailed SR model architectures and introducing extra overhead, which places them in a different category than \textit{FedVSR}. To better understand how our model compares, we tested FedSR with VRT under both heterogeneity settings. Since the original paper did not provide a repository, we implemented the code as faithfully as possible. The results are shown in Table~\ref{tab:extra_baseline}.
\begin{table*}[htbp!]
    \setlength{\tabcolsep}{4pt}
    \centering
    \begin{tabular}{|c|ccc|ccc|ccc|ccc|}
        \hline
        \multirow{2}{*}{\textbf{Algorithm}} 
        & \multicolumn{6}{c|}{\textbf{Less Heterogeneous}}
        & \multicolumn{6}{c|}{\textbf{More Heterogeneous}} \\
        \cline{2-13}
        & \multicolumn{3}{c|}{\textbf{REDS} \cite{Nah_2019_CVPR_Workshops}}
        & \multicolumn{3}{c|}{\textbf{Vid4} \cite{6549107}}
        & \multicolumn{3}{c|}{\textbf{REDS} \cite{Nah_2019_CVPR_Workshops}}
        & \multicolumn{3}{c|}{\textbf{Vid4} \cite{6549107}} \\
        \cline{2-13}
        & {\fontsize{6}{7}\selectfont PSNR}
        & {\fontsize{6}{7}\selectfont SSIM}
        & {\fontsize{6}{7}\selectfont LPIPS}
        & {\fontsize{6}{7}\selectfont PSNR}
        & {\fontsize{6}{7}\selectfont SSIM}
        & {\fontsize{6}{7}\selectfont LPIPS}
        & {\fontsize{6}{7}\selectfont PSNR}
        & {\fontsize{6}{7}\selectfont SSIM}
        & {\fontsize{6}{7}\selectfont LPIPS}
        & {\fontsize{6}{7}\selectfont PSNR}
        & {\fontsize{6}{7}\selectfont SSIM}
        & {\fontsize{6}{7}\selectfont LPIPS} \\
        \hline
        FedSR \cite{yang2025fedsr}
        & 29.70 & 0.8518 & 0.1939
        & 24.89 & 0.7676 & 0.2470
        & 28.85 & 0.8297 & 0.2156
        & 24.70 & 0.7611 & 0.2571 \\
        \textit{FedVSR}
        & 29.96 & 0.8595 & 0.1807
        & 24.97 & 0.7745 & 0.2284
        & 29.12 & 0.8371 & 0.2010
        & 24.78 & 0.7646 & 0.2378 \\
        \hline
    \end{tabular}
    \caption{Quantitative comparison of FedSR \cite{yang2025fedsr}, and \textit{FedVSR} under different heterogeneity settings. Metrics include PSNR, SSIM, and LPIPS for REDS and VID4 datasets.}
    \label{tab:extra_baseline}
\end{table*}

\subsubsection{\textbf{Robustness to Communication Failures}}
\begin{figure*}[htbp!]
    \includegraphics[width=\linewidth]{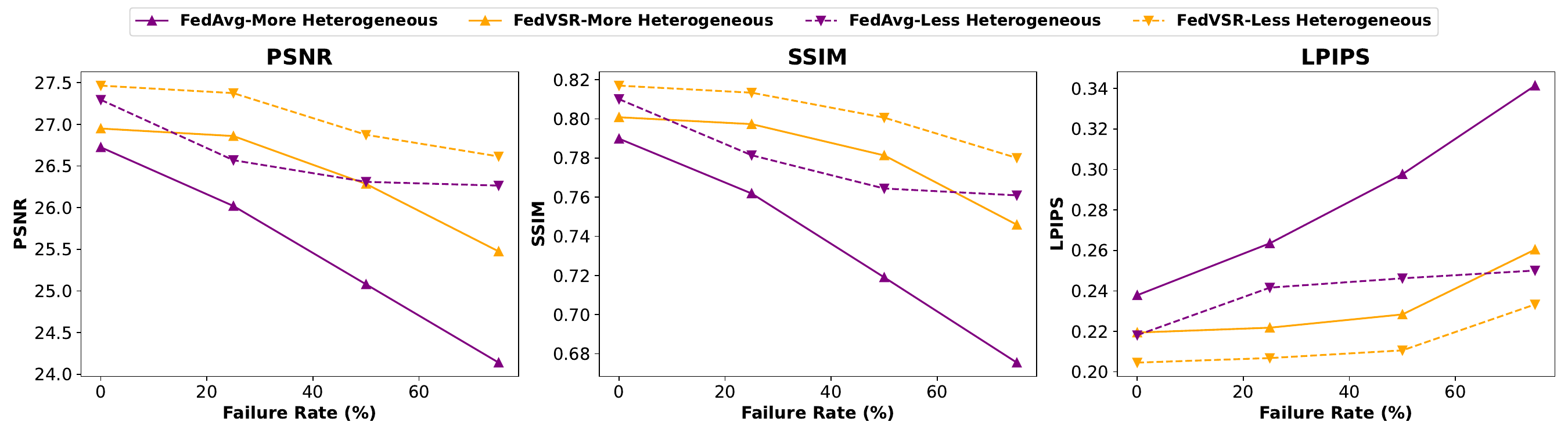}
  \caption{\textit{FedVSR} vs. FedAvg under client upload failures (0–75\%), showing \textit{FedVSR}’s higher robustness.}

  \label{fig:fail}
\end{figure*}

To evaluate the robustness of our method under unreliable communication, we conduct a communication reliability test by simulating scenarios in which a subset of clients fail to upload their updated weights to the server. Specifically, we consider failure rates of 25\%, 50\%, and 75\%, meaning that in each round, the corresponding fraction of clients’ updates are discarded. We test this setting using two models \textit{FedVSR}, and FedAvg. This setup allows us to compare how each approach handles upload failures and missing updates, and how well they maintain performance under increasingly adverse communication conditions. Fig. ~\ref{fig:fail} illustrates the resulting performance for each method across the different failure rates. Our results indicate that \textit{FedVSR} demonstrates greater resilience, maintaining performance even at high failure rates, whereas FedAvg shows significant degradation as the failure rate increases. These findings highlight the advantage of our approach in real-world scenarios where client-server communication may be unreliable.

\subsubsection{\textbf{Client Population Stress Test}}
\begin{figure*}[htb!]
    \includegraphics[width=\linewidth]{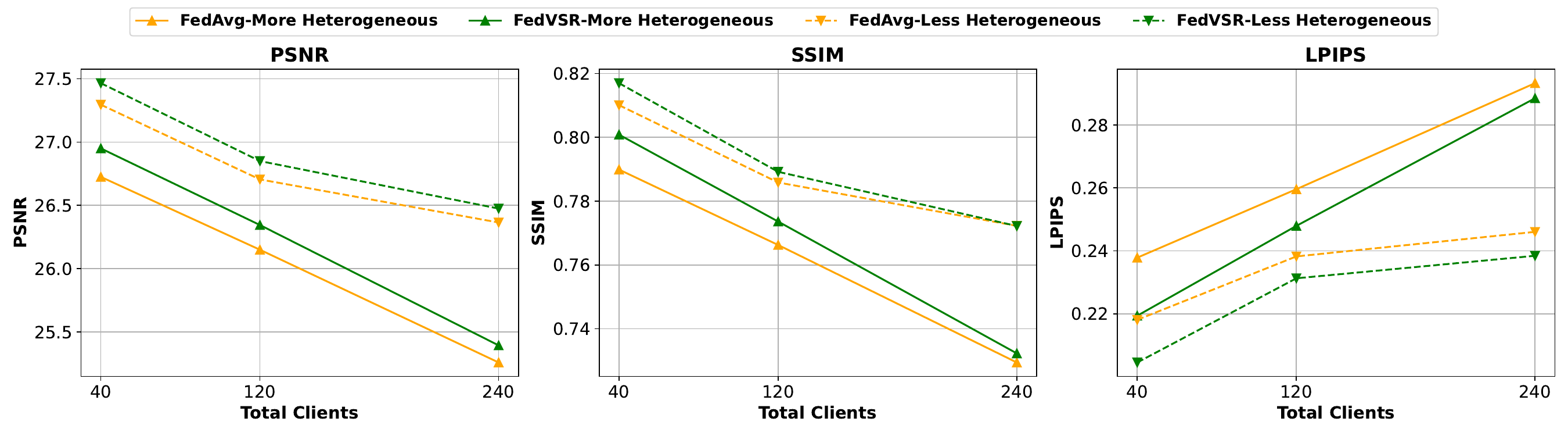}
  \caption{\textit{FedVSR} vs. FedAvg under client population stress test.}

  \label{fig:pop}
\end{figure*}

To evaluate the scalability of our method, we conduct a client population stress test in which we increase the number of clients by three and six times compared to the main experiment~\ref{sec:mainexp}, while keeping the client participation rate constant (10\%). We compare the performance of \textit{FedVSR} and FedAvg under these conditions. This experimental setup allows us to investigate whether our method can maintain high performance when faced with a substantially larger client population, without requiring changes to participation strategies. Fig.~\ref{fig:pop} shows the results of this test, indicating that \textit{FedVSR} consistently outperforms FedAvg even as the number of clients increases, demonstrating the scalability and robustness of our approach in large-scale federated learning scenarios.

\begin{table*}[htbp!]
    \setlength{\tabcolsep}{14pt} 
    \centering
    \begin{tabular}{|c|c|c|ccc|ccc|}
        \hline
        \multirow{2}{*}{\textbf{TC}} & \multirow{2}{*}{\textbf{PR}} & \multirow{2}{*}{\textbf{Algorithm}}  
        & \multicolumn{3}{c|}{\textbf{REDS} \cite{Nah_2019_CVPR_Workshops}} 
        & \multicolumn{3}{c|}{\textbf{Vid4} \cite{6549107}} \\
        \cline{4-9}
        & & & {\fontsize{6}{7}\selectfont PSNR} & {\fontsize{6}{7}\selectfont SSIM} & {\fontsize{6}{7}\selectfont LPIPS}
          & {\fontsize{6}{7}\selectfont PSNR} & {\fontsize{6}{7}\selectfont SSIM} & {\fontsize{6}{7}\selectfont LPIPS} \\
        \hline
        \multicolumn{9}{|c|}{\textbf{Less Heterogeneous}} \\
        \hline
        \multirow{2}{*}{120} & \multirow{2}{*}{3.3\%} & FedAvg & 29.21 & 0.8261 & 0.2136 & 24.36 & 0.7362 & 0.2716 \\
                              &                         & \textit{FedVSR} & 29.36 & 0.8378 & 0.1920 & 24.47 & 0.7472 & 0.2431 \\ [2pt]
        \multirow{2}{*}{240} & \multirow{2}{*}{1.7\%} & FedAvg & 28.23 & 0.8027 & 0.2327 & 23.94 & 0.7079 & 0.2893 \\
                              &                         & \textit{FedVSR} & 28.59 & 0.8155 & 0.2093 & 24.10 & 0.7225 & 0.2562 \\ [2pt]
        \hline
        \multicolumn{9}{|c|}{\textbf{More Heterogeneous}} \\
        \hline
        \multirow{2}{*}{120} & \multirow{2}{*}{3.3\%} & FedAvg & 28.36 & 0.8028 & 0.2453 & 24.09 & 0.7205 & 0.2839 \\
                              &                         & \textit{FedVSR} & 28.54 & 0.8160 & 0.2136 & 24.28 & 0.7376 & 0.2531 \\ [2pt]
        \multirow{2}{*}{240} & \multirow{2}{*}{1.7\%} & FedAvg & 26.95 & 0.7562 & 0.2979 & 23.18 & 0.6641 & 0.3358 \\
                              &                         & \textit{FedVSR} & 27.23 & 0.7742 & 0.2474 & 23.44 & 0.6881 & 0.2839 \\ [2pt]
        \hline
    \end{tabular}
    \caption{Extreme test results for FedAvg and \textit{FedVSR} under different heterogeneity settings. Here, TC denotes the total number of clients and PR denotes the participation rate.}
    \label{tab:extreme}
\end{table*}

\subsubsection{\textbf{Extreme Case: Large Client Population with Low Participation}}
To investigate the behavior of our method under extreme conditions, we conduct a test combining a substantially increased client population with very low participation rates. Specifically, we evaluate scenarios with three and six times more clients than the main experiment~\ref{sec:mainexp}, while reducing the participation rate to approximately one-third (3.3\%) and less than one-fifth (1.7\%) of the original rate (baseline 10\%). We compare the performance of \textit{FedVSR} and FedAvg under these conditions to assess how well each method handles extreme sparsity in client updates combined with a large-scale client population. Table~\ref{tab:extreme} presents the results, demonstrating that \textit{FedVSR} maintains superior performance and robustness even in these challenging scenarios, whereas FedAvg experiences significant performance degradation. These findings highlight the resilience and scalability of \textit{FedVSR} under highly adverse federated learning conditions.

\subsubsection{\textbf{Effect of Local Epochs on Communication-Efficient Performance}}
\begin{figure*}[ht]
    \includegraphics[width=\linewidth]{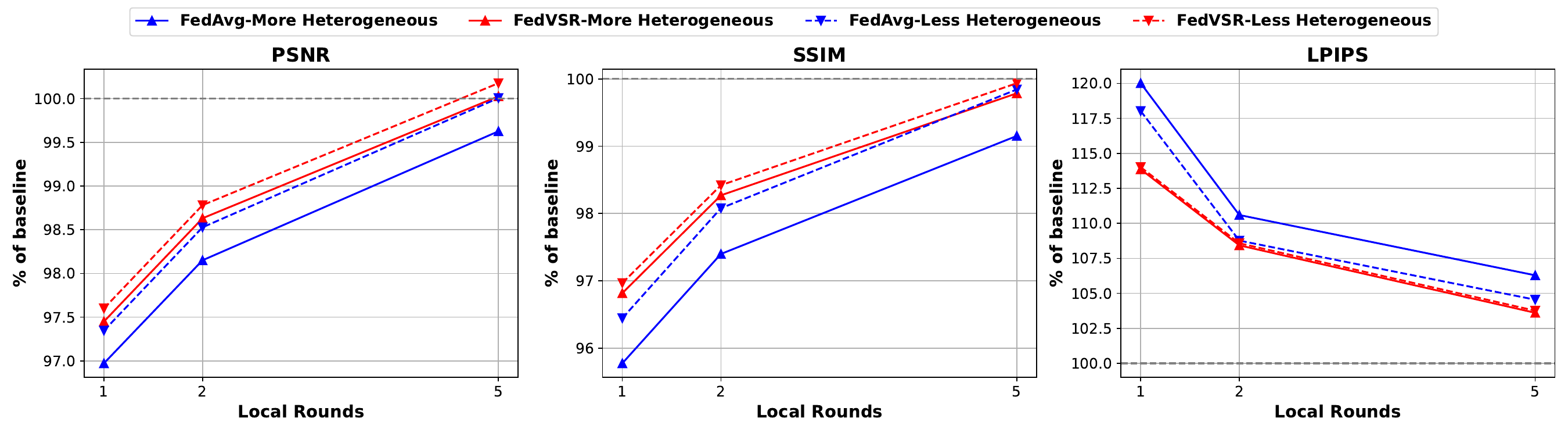}
  \caption{Effect of local epochs on \textit{FedVSR} and FedAvg (\% of 100-round \textit{FedVSR} with 1 local epoch).}

  \label{fig:local}
\end{figure*}

To investigate the trade-off between local computation and communication efficiency, we compare the performance of \textit{FedVSR} and FedAvg after 20 global rounds using different numbers of local epochs per client. These experiments are measured relative to our main experiment, which consists of 100 global rounds with 1 local epoch per client. The results are presented in the Fig. ~\ref{fig:local} as percentages of the main experiment’s performance. This comparison allows us to assess whether increasing the number of local epochs can compensate for fewer global rounds while highlighting the differences between \textit{FedVSR} and FedAvg under limited communication.

\subsubsection{\textbf{Resource and Efficiency Overhead}}
\label{sec:overhead}
Table~\ref{tab:overhead_phases} reports the computation and memory overhead of different FL algorithms relative to FedAvg. We select FedAvg as the baseline because it is a vanilla FL algorithm that introduces no additional overhead. To ensure accuracy, each measurement was repeated multiple times. We also report results as relative percentages, since our experiments involved three different VSR models with varying computational requirements. A value of 0\% indicates no overhead beyond FedAvg, while higher values indicate additional computational or memory costs. Presenting the results in this way provides a clearer comparison across models. For these measurements, we assume an ideal network, as the communication cost of transmitting parameters is negligible compared to training time, as discussed earlier~\ref{sec:repro}. If we were to consider communication cost, in \textit{FedVSR} each client only needs to send a single scalar, unlike SCAFFOLD or FedSR, which incur a much heavier overhead. As shown, \textit{FedVSR} not only delivers strong performance but also incurs minimal overhead compared to its counterparts. 

\begin{table*}[htbp!]
    \setlength{\tabcolsep}{16pt}
    \centering
    \begin{tabular}{|c|c|c|c|c|c|}
        \hline
        \textbf{Model} & \textbf{Phase} & \textbf{Latency} & \textbf{Energy} & \textbf{GPU Mem} & \textbf{GPU Util} \\
        \hline
        \multirow{2}{*}{\textit{FedVSR}} 
            & Training     & 0.25  & 0.73  & \multirow{2}{*}{0.01} & \multirow{2}{*}{0.17} \\
            & Aggregation & 5.39  & 4.36  &  &  \\
        \hline
        \multirow{2}{*}{SCAFFOLD} 
            & Training     & 10.56 & 11.35 & \multirow{2}{*}{5.53} & \multirow{2}{*}{4.85} \\
            & Aggregation & 17.44 & 21.12 &  &  \\
        \hline
        \multirow{2}{*}{FedProx} 
            & Training     & 17.32 & 19.82 & \multirow{2}{*}{0.71} & \multirow{2}{*}{2.93} \\
            & Aggregation & 39.74 & 39.20 &  &  \\
        \hline
        \multirow{2}{*}{FedSR} 
            & Training     & 1.90  & 4.61  & \multirow{2}{*}{3.16} & \multirow{2}{*}{11.92} \\
            & Aggregation & 314.62 & 620.42 &  &  \\
        \hline
        \multirow{2}{*}{FedMedian} 
            & Training     & 0  & 0  & \multirow{2}{*}{0} & \multirow{2}{*}{0} \\
            & Aggregation & 2.15  & 1.51  &  &  \\
        \hline
    \end{tabular}
    \caption{Computation and memory overhead of different FL methods relative to FedAvg (\%).GPU memory and utilization are reported for both Training and Aggregation phases.}
    \label{tab:overhead_phases}
\end{table*}

\subsection{Ablation Study}
\label{sec:ablation}
\subsubsection{\textbf{Evaluating High Frequency Loss on Centralized Setting}} 
\label{sec:abl_loss_center}
The results clearly demonstrate that adding $\mathcal{L}_{\text{HiFr}}$ to centralized training negatively impacts performance across all metrics. This degradation is observed consistently in both the less and more heterogeneous data splits. This outcome confirms that the $\mathcal{L}_{\text{HiFr}}$ loss is not a generic VSR improvement. Rather, it is a necessary intervention specifically designed to counteract the core failings of Federated Learning in low-level vision tasks. In centralized training, the model has sufficient, high-bandwidth data access to efficiently capture high-frequency details using the standard VSR loss. Introducing the DWT-based loss in this setting acts as a redundant constraint, likely leading to over-regularization or gradient noise, which harms the final model quality. Our finding establishes that $\mathcal{L}_{\text{HiFr}}$ is a crucial, non-trivial component of the $\textit{FedVSR}$ framework designed exclusively to fill the performance gap created by the inherent constraints of federated learning for VSR.
\begin{table*}[htbp!]
    \setlength{\tabcolsep}{18pt} 
    \centering
    \begin{tabular}{|c|ccc|ccc|}
        \hline
        \multirow{2}{*}{\textbf{Algorithm}}  
        & \multicolumn{3}{c|}{\textbf{REDS} \cite{Nah_2019_CVPR_Workshops}} 
        & \multicolumn{3}{c|}{\textbf{Vid4} \cite{6549107}} \\
        \cline{2-7}
        & {\fontsize{6}{7}\selectfont PSNR} & {\fontsize{6}{7}\selectfont SSIM} & {\fontsize{6}{7}\selectfont LPIPS}
        & {\fontsize{6}{7}\selectfont PSNR} & {\fontsize{6}{7}\selectfont SSIM} & {\fontsize{6}{7}\selectfont LPIPS} \\
        \hline
        \multicolumn{7}{|c|}{\textbf{Less Heterogeneous}} \\
        \hline

        VRT    & 32.20 & 0.9007 & 0.1342 & 26.31 & 0.8266 & 0.1780 \\
        VRT+\(\mathcal{L}_{\text{HiFr}}\)  & 31.11 & 0.8702 & 0.1543 & 25.27 & 0.8020 & 0.2064 \\
        \hline
        \multicolumn{7}{|c|}{\textbf{More Heterogeneous}} \\
        \hline
    
        VRT  & 31.06 & 0.8688 & 0.1505 & 25.25 & 0.7900 & 0.1993 \\
        VRT+\(\mathcal{L}_{\text{HiFr}}\)   & 30.12 & 0.8424 & 0.1700 & 24.50 & 0.7666 & 0.2238 \\
        \hline
    \end{tabular}
    \caption{Ablation study comparing VRT performance with and without the $\mathcal{L}_{\text{HiFr}}$ loss under centralized training. }
    \label{tab:abl_loss_center}
\end{table*}

\subsubsection{\textbf{Evaluating High Frequency Loss and Adaptive Aggregation}} To further analyze the impact of the DWT loss term and loss-aware adaptive aggregation, we conducted a series of experiments. Using FedAvg as a baseline, we added each component separately to evaluate their individual effects. To assess the impact of the \(\mathcal{L}_{\text{HiFr}}\) term, we applied it to FedAvg and observed the changes (F-\(\mathcal{L}_{\text{HiFr}}\)). Similarly, to evaluate loss-aware adaptive aggregation, we first trained VSR models using a greedy aggregation strategy (F-GLA) that prioritizes low-loss updates in all rounds. We then compared this approach to our adaptive aggregation (F-ALA). As shown in Table~\ref{tab:ablation_study}, the DWT loss term consistently outperforms loss-aware adaptive aggregation in all cases. Additionally, both methods individually surpass FedAvg. However, in some cases, the greedy aggregation strategy resulted in only marginal improvements or even a drop in overall performance especially in "more heterogeneous" setting based on PSNR and SSIM metrics. These findings further highlight the importance of high-frequency information not only for improving overall performance but also as a reliable metric for identifying valuable client weights.

\begin{table*}[htbp!]
    \setlength{\tabcolsep}{11pt} 
    \centering
    \begin{tabular}{|c|c|cc|cc||cc|cc|}
        \hline
        \multirow{3}{*}{\makecell[c]{\centering \textbf{Model}}} & \multirow{3}{*}{\textbf{Setting}}  
        & \multicolumn{4}{c||}{\textbf{Less Heterogeneous}} 
        & \multicolumn{4}{c|}{\textbf{More Heterogeneous}} \\
        \cline{3-10}
        & & \multicolumn{2}{c|}{\textbf{REDS} \cite{Nah_2019_CVPR_Workshops}} & \multicolumn{2}{c||}{\textbf{Vid4} \cite{6549107}}  
          & \multicolumn{2}{c|}{\textbf{REDS} \cite{Nah_2019_CVPR_Workshops}} & \multicolumn{2}{c|}{\textbf{Vid4} \cite{6549107}} \\
        & & {\fontsize{6}{7}\selectfont PSNR} & {\fontsize{6}{7}\selectfont SSIM} & {\fontsize{6}{7}\selectfont PSNR} & {\fontsize{6}{7}\selectfont SSIM}
          & {\fontsize{6}{7}\selectfont PSNR} & {\fontsize{6}{7}\selectfont SSIM} & {\fontsize{6}{7}\selectfont PSNR} & {\fontsize{6}{7}\selectfont SSIM} \\
        \hline
        \multirow{2}{*}{\makecell{\centering \textbf{VRT}}} 
            & F-\(\mathcal{L}_{\text{HiFr}}\) \rule{0pt}{3ex}   & 29.80 & 0.8543 & 24.91 & 0.7681 & 28.98 & 0.8322 & 24.70 & 0.7577 \\
            & F-ALA \rule{0pt}{2ex}  & 29.75 & 0.8531 & 24.89 & 0.7681 & 28.90 & 0.8313 & 24.64 & 0.7559 \\ [2pt]
            & F-GLA \rule{0pt}{2ex}  & 29.70 & 0.8523 & 24.86 & 0.7670 & 28.83 & 0.8279 & 24.55 & 0.7500 \\ [2pt]
        \hline
        \multirow{2}{*}{\makecell{\centering \textbf{RVRT}}} 
            & F-\(\mathcal{L}_{\text{HiFr}}\) \rule{0pt}{3ex} & 31.42  & 0.8888 & 26.02  & 0.8178 & 30.24  & 0.8617 & 25.84 & 0.8042 \\
            & F-ALA \rule{0pt}{2ex} & 31.38  & 0.8880 & 26.00  & 0.8171 & 30.19  & 0.8601 & 25.79 & 0.8041 \\ [2pt]
          & F-GLA \rule{0pt}{2ex}  & 31.31 & 0.8879 & 25.95 & 0.8155 & 29.88 & 0.8576 & 25.60 & 0.8032 \\ [2pt]
        \hline
        \multirow{2}{*}{\makecell{\centering \textbf{IART}}}
            & F-\(\mathcal{L}_{\text{HiFr}}\) \rule{0pt}{3ex} & 30.79  & 0.8801 & 25.46 & 0.8001 & 29.64 & 0.8508 & 25.21 & 0.7837 \\
            & F-ALA \rule{0pt}{2ex} & 30.70  & 0.8789 & 25.44 & 0.7998 & 29.60 & 0.8500 & 25.18 & 0.7831 \\ [2pt]
          & F-GLA \rule{0pt}{2ex}  & 30.57 & 0.8730 & 25.42 & 0.7951 & 29.50 & 0.8482 & 24.99 & 0.7761 \\ [2pt]
        \hline
    \end{tabular}
    \caption{Ablation study on the impact of \(\mathcal{L}_{\text{HiFr}}\) and adaptive aggregation under different heterogeneity settings.}
    \label{tab:ablation_study}
\end{table*}

\section{Conclusion}

\label{sec:conclusion}
In this work, we introduced \textit{FedVSR}, the first FL framework specifically designed for VSR, addressing privacy concerns while preserving high-quality performance. Unlike existing FL approaches that struggle with low-level vision tasks, our model-agnostic and stateless design ensures adaptability across various VSR architectures without prior model knowledge or historical weight storage. By incorporating a lightweight loss term, we effectively guide local updates and enhance global aggregation, resulting in notable performance improvements over general FL baselines. Our method functions as a plug-and-play solution, and extensive experiments confirm its superiority over existing FL approaches, demonstrating its potential for real-world, privacy-sensitive applications. Furthermore, our study highlights key challenges in applying FL to low-level vision tasks, offering valuable insights for future research in this under-explored domain. As the first FL-based approach for VSR, this work establishes a strong baseline and provides a clear research path for future advancements. Despite promising results, a gap remains between FL-based VSR and centralized approaches. Future work could improve aggregation with task-specific strategies for low-level vision and enhance our loss-guided optimization for greater stability and effectiveness in FL-based VSR.

\paragraph{\textbf{Future Work}} A key direction for future work is extending $\textit{FedVSR}$ to handle higher-resolution videos and scenarios where clients may contribute data at differing native resolutions.

{
    \small
    \bibliographystyle{ieeenat_fullname}
    \bibliography{main}
}


\end{document}